# CLINIQA: A Machine Intelligence Based Clinical Question Answering System

M. A. H. Zahid, Ankush Mittal, R.C. Joshi, and G. Atluri

*Abstract*-- The recent developments in the field of biomedicine have made large volumes of biomedical literature available to the medical practitioners. Due to the large size and lack of efficient searching strategies, medical practitioners struggle to obtain necessary information available in the biomedical literature. Moreover, the most sophisticated search engines of age are not intelligent enough to interpret the clinicians' questions. These facts reflect the urgent need of an information retrieval system that accepts the queries from medical practitioners' in natural language and returns the answers quickly and efficiently. In this paper, we present an implementation of a machine intelligence based CLINIcal Question Answering system (CLINIQA) to answer medical practitioner's questions. The system was rigorously evaluated on different text mining algorithms and the best components for the system were selected. The system makes use of Unified Medical Language System for semantic analysis of both questions and medical documents. In addition, the system employs supervised machine learning algorithms for classification of the documents, identifying the focus of the question and answer selection. Effective domain-specific heuristics are designed for answer ranking. The performance evaluation on hundred clinical questions shows the effectiveness of our approach.

*Index Terms*-- Biomedical, Question Answering Systems, Text mining, and Semantic analysis.

## I. INTRODUCTION

THE development in the fields of biomedicine, electronic publishing, and computing technology has led to the rapid growth of the biomedical literature available online to the medical practitioners. The growth in biomedical literature can be assessed from the statistics about PUBMED database of National Library of Medicine (NLM), which consists of more than 14 million citations. The statistics also reveal that every year four hundred thousands new citations are being added and over 120 million searches are being conducted [1]. Due to the large size of information and lack of efficient searching strategies, most of the times medical practitioners and biologists fail to locate the desired information available in the literature. There are two major disadvantages of the most sophisticated search engines of the age: (1) search engines retrieves documents that contain high frequency of keywords in the query, and often the semantics are neglected, (2) the search results in a set of documents and user needs to navigate through them to locate the required information which is tedious and time consuming. Generic search engines seldom have any advantage in specialized domains like biomedicine because the emphasis of search engines is more on individual keywords rather on the phrases, as desired. This reflects the need of a system that can intelligently interpret the clinicians' questions and retrieve the appropriate snippets from the document collection.

Evidence-based medicine has emerged as a widely accepted paradigm for medical practice, which emphasizes the importance of evidence from patient-centered clinical research in the health care process. It refers to the use of evidence from scientific and medical research in making decisions related to the care of an individual patient. Therefore, medical practitioners are urged to practice evidence based medicine when faced with the questions about patient care [11,12]. Due to the pressurized atmosphere of busy practice, clinicians more likely seek answers from readily available sources [13]. Although there is a large amount of information available to the clinicians through different sources such as digital libraries, electronic journal, and WWW, they often face difficulties in finding answers to their questions about a specific clinical problem [14].

A study revealed that the average time required in obtaining an adequate answer to the physician's question ranged from 2.4 to 6.5 minutes [15]. In another study conducted with a test set of 100 medical questions collected from medical students in a specialized domain, a thorough search in Google was unable to obtain relevant documents within top five hits for 40% of the questions [16]. Moreover, due to busy practice schedules physicians spend less than 2 minutes

R. C. Joshi (e-mail: joshifcc@iitr.ernet.in) G. Atluri (e-mail: atlurpec@iitr.ernet.in).

on average seeking an answer to a question. Thus, most of the clinical questions remain unanswered [17]. The aforementioned studies clearly indicate the urgent need of a system which answers clinicians' questions effectively and quickly related to patient care.

The existing information retrieval systems usually fail to provide satisfactory results due to domain specific nature of the biomedical text, which comprises of complex technical terms and their inconsistent use in the domain [8]. For example, there are a large number of long multiword expressions such as "Melanogaster 5`-phosphoribosylaminoimidazole carboxylase-5'-phosphoribosyl-4-(N-succinocarboxamide)-5-aminoimidazole synthetase (Ade5)mRNA, complete cds". Moreover, the term order variations and abbreviations are common in this field [9]. Beside these domain-specific characteristics, there are many challenges with respect to question answering, which form a clear demarcation between biomedical and open domain question answering systems (OQAS). The initial feasibility study on medical question answering systems, conducted by Nui et. al. [10], gives a good analysis of the limitations of OQAS techniques when applied to Medical Question Answering Systems (MQAS).

In this paper, we present an implementation of the intelligent Clinical Question Answering system developed for medical practitioners and for the users acquainted with medical terminology. Currently, the system works on the pancreatic cancer abstracts collected from PUBMED, but the architecture is robust and applicable to any disease and knowledge source. The collected documents are classified and indexed based on the domain requirements. Similarly, the system classifies the clinical questions to obtain question focus and expected answer types. Simple yet effective strategies are developed for answer extraction and ranking. Finally, the answer based on the rank is returned to the user.

The paper is organized as follows, in section II, we give a detailed taxonomy of CLINIQA,

where each component is explained and analyzed in detail. In section III, the system is evaluated on user effort and Mean reciprocal Rank. A survey of related work is presented in section IV.

## II. CLINIQA: System Architecture

There are four major components in CLINIQA, which are question classification, query formulation, answer extraction, and answer ranking. The architecture is shown in Fig. 1.

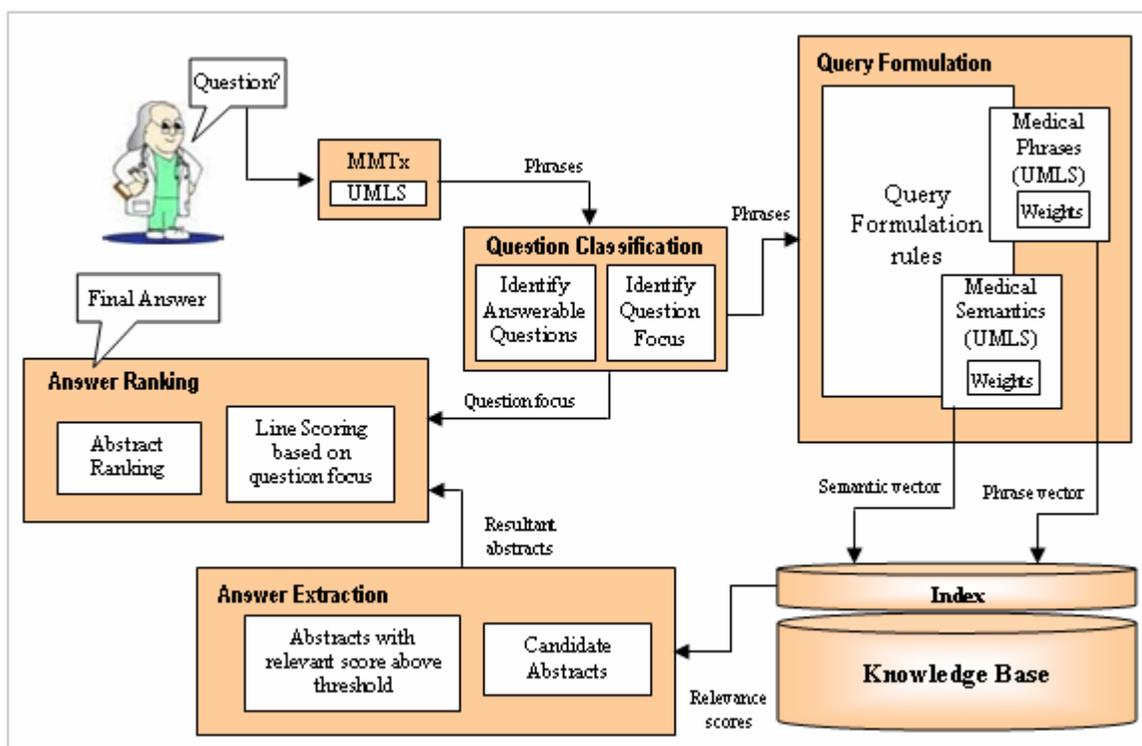

Fig. 1. CLINIQA system architecture.

Classification and indexing of the documents are performed as a data preprocessing step, which significantly improve the retrieval performance, and are explained in section *A*. CLINIQA accepts natural language questions and parses them using MMTx [6], a programming implementation of MetaMap [25], which maps free text to their Unified Medical Language System (UMLS) [2] concepts and related semantic types. The obtained medical phrases and its semantic types are then given as input to the question classification and query formulation modules. The question classification uses MMTx output to classify the user question into a predefined set of classes, which is discussed in section *B*. The query formulation module also uses the result of MMTx and assigns weights to the medical phrases and semantics. These weights are then depicted as a vector of phrases and their semantics. The query formulation module is discussed in section *C*. The vectors generated by the query formulation module are then used along with the document index to find the most relevant documents based on similarity. The answer extraction module selects the documents that have relevance score or cosine value above a certain threshold, as discussed in section *D*. Finally, the answer ranking module ranks the document and assigns scores to the sentences based on the co-occurrence of question terms and the required answer type. The answer ranking module is discussed in section *E*.

*A. Data Preprocessing*

   *1) Biomedical document indexing*

To answer the user's questions quickly and effectively using a knowledge-base that scales to gigabytes of text, the data must be preprocessed and each document is to be represented by a set of keywords that describe the document. The unique property of the biomedical corpus is that, it contains a large set of multiword expressions which are out of the scope of general English language resources. Many methods have been developed for automatic indexing of biomedical documents [9,25]. Keeping in view the unique property of the biomedical corpus mentioned above, we index the documents based on both medical phrases and semantics. The semantic indexing is useful in interpreting the content of the document. We used the MMTx program, which first parses text and then separates it into noun phrases. Each noun phrase is then mapped into UMLS concepts and each concept is assigned some weight where the highest weight represents the most likely concept the document represents. MMTx also assigns a semantic tag to each of the concepts. An example of the indexing used in the system is shown in Table I.

TABLE I
A SAMPLE ABSTRACT WITH MEDICAL PHRASE INDEX AND SEMANTIC TAGS

| |
|---|
| PMID: 16169155 |
| Title: "Mutant KRAS in the initiation of pancreatic cancer." |
| *Abstract:*<br>Pancreatic ductal adenocarcinoma is the most common pancreatic neoplasm. There are approximately 33,000 new cases of pancreatic ductal adenocarcinoma annually in the United States with approximately the same number of deaths. Surgery represents the only opportunity for cure, but this is restricted to early stage pancreatic cancer. Pancreatic ductal adenocarcinoma evolves from a progressive cascade of cellular, morphological and architectural changes from normal ductal epithelium through preneoplastic lesions termed pancreatic intraepithelial neoplasia (PanIN). These PanIN lesions are in turn associated with somatic alterations in canonical oncogenes and tumor suppressor genes. Most notably, early PanIN lesions and almost all pancreatic ductal adenocarcinomas involve mutations in the K-ras oncogene. Thus, it is believed that activating K-ras mutations are critical for initiation of pancreatic ductal carcinogenesis. This has been proven through elegant genetically engineered mouse models in which a Cre-activated K-Ras(G12D) allele is knocked into the endogenous K-Ras locus and crossed with mice expressing Cre recombinase in pancreatic tissue. As a result, mechanistic insights are now possible into how K-Ras contributes to pancreatic ductal carcinogenesis, what cooperating events are required, and armed with this knowledge, new therapeutic approaches can be pursued and tested. |
| *Index: medical phrases*<br>pancreatic adenocarcinoma; duct; most; common; neoplasm, pancreatic; approximately; new; cases; annually; united states; same; number; deaths; surgery; restricted; early; stage pancreatic cancer; evolving; progressive; cellularmorphology;architecture;changed;normal;epithelium;lesions;intraepithelial neoplasm; turns; association; soma;altered; oncogenes; genes, tumor suppressor; notes; pancreatic duct; adenocarcinomas; involved; mutations; k-ras oncogene; belief; activate; kras; crises; initiation; carcinogenesis; proven; genetic engineering; mouse; models; allele; endogenous; crossed;mice; expressor; cre recombinase; tissue; result; mechanism; insight; possible; cooperation; events; arm; knowledge; therapeutic; approaches; testing |
| *Index: semantic tags*<br>Neoplastic Process; Body Part, Organ, or Organ Component; Quantitative Concept; Functional Concept, Qualitative Concept; Temporal Concept; Geographic Area; Organism Function; Therapeutic or Preventive Procedure; Diagnostic Procedure; Anatomical Structure; Occupation or Discipline; Finding; Tissue; Mental Process; Cell Component; Gene or Genome; Intellectual Product; Genetic Function; Idea or Concept; Molecular Biology Research Technique; Mammal; Intellectual Product, Research Device; Spatial Concept; Medical Device; Amino Acid, Peptide, or Protein, Enzyme; Social Behavior; Event; Body Location or Region; Research Activity. |

Fig. 2 (a) shows the medical term frequency for a small set of medical documents, where the density of the graph represents the frequency of occurrence of the words in the documents. The terms which are common in all the documents are not useful for indexing. Fig. 2(a) and Fig. 2(b) show the frequency of medical phrases and semantic tag in a sample of documents collected from the corpus.

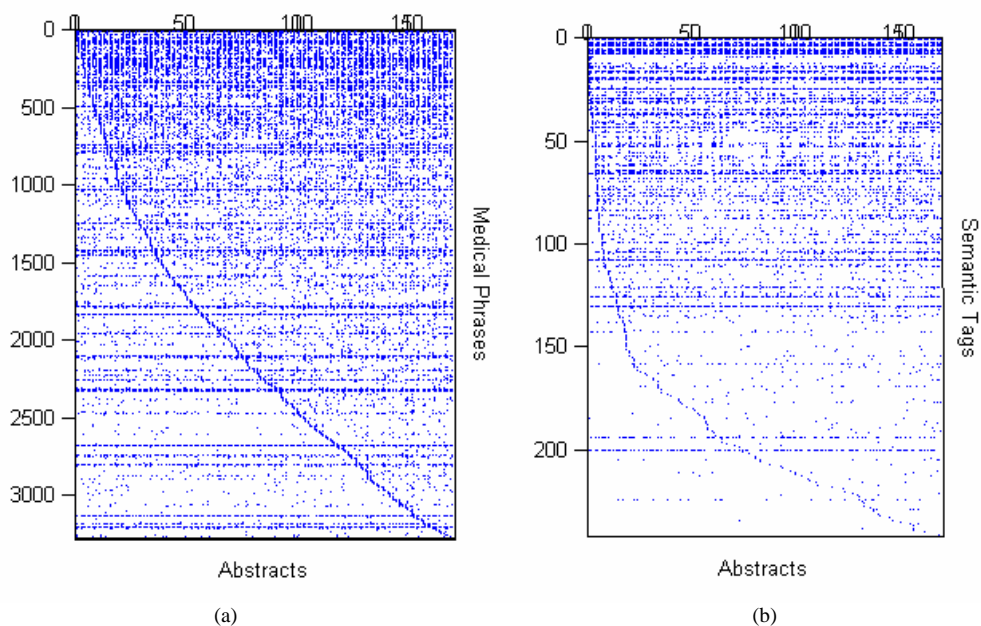

(a) (b)
Fig. 2. (a) Medical phrases × Abstracts (b) Semantic Tags × Abstracts

In this phase documents are indexed using unique medical phrases and semantic tags. The efficiency of the indexing is inversely proportional to the document length. As the length of document increases the size of the index will decrease. In a sample of 1700 abstracts, 9260 unique phases and 345 unique medical concepts are identified.

*2) Term weighing and scoring*

The system is assessed after assigning weights to each term in the document based on different weighing schemes. The weights are assigned to the terms based on the statistical properties of the terms in the documents. For example, *Term Frequency* (TF) is the measure of how many times a term has appeared in the document. The TF weighing scheme considers all the terms equally important, but this is always not true. Certain terms have no discriminating power but have more occurrences in the documents. The proper strategy to achieve this is to assign less weights to frequently appearing terms and high weights to rarely appearing terms. *Inverse Document Frequency* (IDF) follows this weighing scheme and can be measured as follows:

$$IDF = \log(\frac{N}{DF}), \text{ where } N \text{ is the number of documents in the text collection,}$$
$$DF \text{ is the number of documents in which the term appeared}$$

Long documents generally contain certain terms repeated more frequently, therefore the TF may be large for long documents than the shorter ones. As a result long documents pretend to be more relevant to the words that occur more frequently, though it may not be the case in reality. Both TF and IDF suffer with this drawback. To compensate this effect, normalization is used, which is calculated as follows:

$$Normalized\ term\ frequency = \frac{TF \times IDF}{Euclidean\ length\ of\ document\ vector}$$

$$Euclidean\ length\ of\ TF \times IDF = \sqrt{w_1^2 + w_2^2 + .... + w_n^2}$$
$$\text{where } w_i \text{ is the } TF \times IDF \text{ weight of the } i^{th} \text{ term in the document}$$

To illustrate this, consider the three graphs shown in Fig. 3, representing the Retrieval Status Value (RSV) of each document. The relevant documents for an example query are identified as 2, 59, 91, and 168 by domain experts. However, when TF and IDF are used the document 93 gives RSV more than the other relevant documents as shown in Fig. 3(a) and Fig. 3(b) respectively. On the other hand, when normalization is used, the relevant documents have higher RSV as shown in Fig. 3(c).

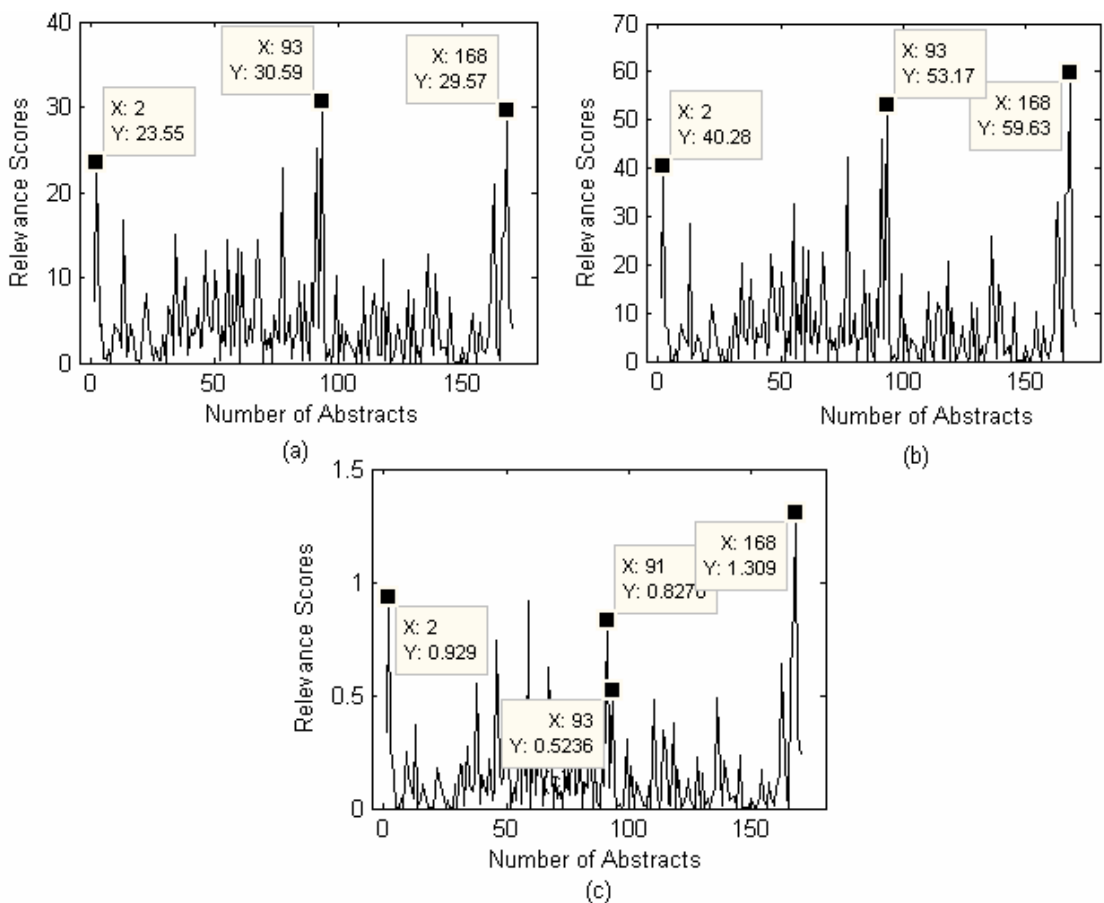

Fig. 3. Retrieval status value values for a sample query using (a) phrase frequency, (b) Inverse Document Frequency and (c) Normalized weights

In this phase three matrices: $Abstracts \times Medical\ Phrases$, $Abstracts \times Semantic\ Tags$ and $Abstracts \times (Medical\ Phrases + Semantic\ Tags)$ are generated. Normalized weights are assigned to each term in the matrices. Scatter graphs depicting a sample of $Abstracts \times Medical\ Phrases$ and $Abstracts \times Semantic\ Tags$ are shown in figure 2.

*3) Document classification*

The questions involving intervention and non intervention are the potential candidates for answering based on evidence from biomedical literature and other medical resources [6]. As only intervention and non-intervention question can be answered, we constrain CLINIQA's knowledgebase to the evidence based medical literature. This leads to efficient utilization of memory, reduces the search space and hence time. To identify the evidence based documents, we used classification techniques to classify the medical documents into three categories, non-evidence, intervention and non-intervention documents. Intervention and non-intervention based documents comprise the evidence based class of documents. Ely et al. [6] gave a hierarchical classification of medical questions as shown in Fig. 4.

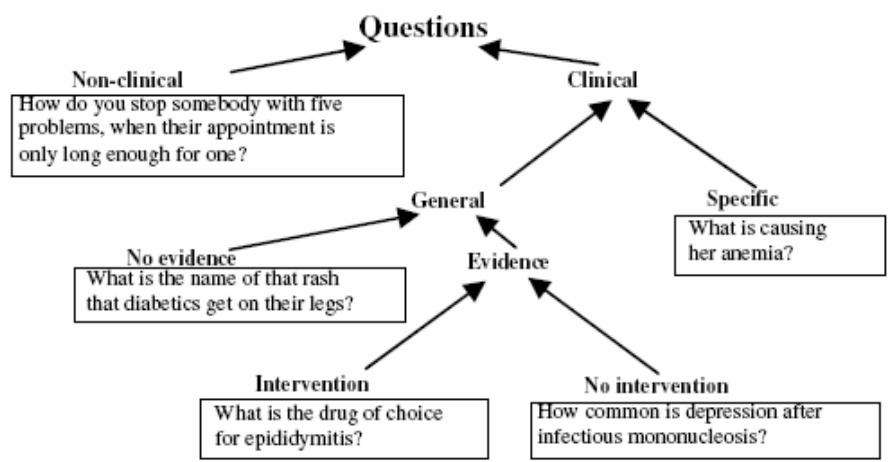

Fig. 4. "Evidence taxonomy" created by Ely et al. [9]

We compared the accuracy of five different classifies: SVM, Naïve Bayes, K-nearest neighbor, Decision Tree based J48, and Fuzzy classifier. A good discussion on these classifiers is given in [28]. A precise discussion of these classifiers is given below:

  *a)    Support Vector Machine*

Support Vector Machine (SVM) is a well established supervised machine learning technique. It relies on the optimal hyperplane algorithm. SVMs minimize the empirical error, complexity of the classifier and capability of learning in high dimensional space with relatively less training samples. If data to be classified by the SVM is linearly separable then a simple linear SVM can be used for classification. On the other hand, if the data cannot be separated by a hyperpalne then the data can be projected into a higher dimensional Hilbert space and can be separated linearly in the higher dimensional projection space.

Let the training pattern be represented as $P = (x, y)$, where $x$ is an element of feature space and $y$ is an element of class space. Let $\Phi$ denote a nonlinear map $\Phi : R^l \rightarrow H$, where H is a higher dimensional Hilbert space and l is the dimensionality of x. SVM construct the optimal separating plane in H. Therefore the decision function is of the form

$$f(x) = \sum_{i=1}^{n} \lambda_i y_i K(x, x_i) + \theta$$

where $x_i$ a learning vector of size n, x is is a test vector, K is a kernel function, and $\lambda_i$ are non-negative lagrange multipliers associated with the quadratic optimization problem:

$$\text{minimize} \quad \frac{1}{2} \mathbf{w}^T \mathbf{w} + D \sum_{i=1}^{n} \xi_i$$

$$\text{subject to} \quad y_i (\mathbf{w}^T \Phi(x_i) + \theta) \geq 1 - \xi_i \text{ for } i = 1, 2, ..., n$$

where w and $\theta$ are the parameters of the optimal separating plane in *H*. Whereas, *D* is a parameter which penalizes the error and $\xi_i$ are positive slack variables.

We used exponential radial bias function (ERBF) kernel, $K(x, x_i) = \exp\{-\gamma |x - x_i|\}$ for classifying the medical documents and questions. We used $D = 500$ and $\gamma = 0.005$, moreover, the selection by cross validation is also considered.

b)   *K Nearest Neighbor*

KNN classifier is an instance-based learning algorithm that is based on a distance function for pairs of observations, such as the Euclidean distance. The Euclidian distance between two points $X = (x_1, x_2,..., x_n)$ and $Y = (y_1, y_2,..., y_n)$ is computed as $d(X,Y) = \sqrt{\sum_{i=1}^{n}(x_i - y_i)^2}$ .

In this classification paradigm, *k*-nearest neighbors of a training data is computed first. Then the similarities of one sample from testing data to the *k*-nearest neighbors are aggregated according to the class of the neighbors, and the testing sample is assigned to the most similar class. One of advantages of KNN is that it is well suited for multi-modal classes as its classification decision is based on a small neighborhood of similar objects. So, even if the target class is multi-modal, it can still lead to good accuracy. We used $k = 3$ for our experiments, this value of *k* gives optimal results on the medical document and questions datasets.

c) Naive Bayes

Naïve Bayes (NB) is a probabilistic classifier. The basic idea in NB approaches is to use the joint probabilities of words and categories to estimate the probabilities of categories given a document. The naïve part of NB methods is the assumption of word independence, i.e., the conditional probability of a word given a category is assumed to be independent from the conditional probabilities of other words given that category. This assumption makes the computation of the NB classifiers far more than the exponential complexity of non-naïve Bayes approaches because it does not use word combinations as predictors.

Suppose there are m classes, $C_1, C_2, \ldots, C_m$. Given an unknown data sample $X$, the classifier will predict that $X$ belongs to the class having highest posterior probability, conditioned on $X$. That is, the naïve Bayesian classifier assigns an unknown sample $X$ to the class $C_i$ if and only if $P(C_i/X) > P(C_j/X)$ for $i \leq j \leq m, j \neq i$. thus we maximize $P(C_i/X)$. The class $C_i$ for which $P(C_i/X)$ is maximized is called maximum posteriori hypothesis. By Bayes Theorem,

13$$P(C_i / X) = \frac{P(X / C_i) P(C_i)}{P(X)}$$

Given a data set with many attributes, it would be computationally expensive to compute $P(X / C_i)$. In order to reduce computation in evaluating $P(X / C_i)$, the naïve assumption of class conditional independence is made. This presumes that the values of the attributes are conditionally independent of one another. Thus

$$P(X / C_i) = \prod_{k=1}^{n} P(x_k / C_i)$$

*d)*      *Decision Tree (J48)*

A decision tree is a simple structure where non-terminal nodes represent tests on one or more attributes and terminal nodes reflect decision outcomes. The basic algorithm for decision tree is greedy and constructs the tree in top-down recursive divide and conquer manner. The basic strategy is as follows [27]:

---

**Basic Strategy for Decision tree:**

- The tree starts as a single node representing the training samples.
- If the samples are all of the same class, then the node becomes a leaf and is labeled with that class.
- Otherwise, the algorithm uses an entropy-based measure known as information gain as a heuristic for selecting the attribute that will best separate the samples into individual classes.
- A branch is created for each known value of the test attribute, and the samples are partitioned accordingly.
- The algorithm uses the same process recursively to form a decision tree for the samples at each partition. Once an attribute has occurred at a node, it need not be considered in any of the node's descendents.
- The recursive partitioning stops only when any one of the following conditions is true:
    1. All samples for a given node belong to the same class.
    2. There are no remaining attributes on which the samples may be further partitioned.
    3. There are no samples for the branch

---

Decision tree algorithms are unstable. Slight variations in the training data can result it different attribute selections at each node within the tree. The effect can be significant since attribute choices affect all descendent subtrees.

*e)   Linear Discriminent Analysis*

Linear Discriminant Analysis (LDA) is used for classification of objects into two or more classes. LDA is closely related to linear regression and Principle Component Analysis (PCA). Let the training pattern are represented as $P = (x, y)$, where $x$ is an element of feature space and $y$ is an element of class space. LDA for two class problem approaches the problem by assuming the probability density functions $P(\vec{x}/y=0)$ and $P(\vec{x}/y=1)$ are normally distributed and with identical covariance $\Sigma_{y=0} = \Sigma_{y=1} = \Sigma$. It can be shown that the required probability $P(y/\vec{x})$ depends only on the dot product $\vec{w}.\vec{x}$ where

$$\vec{w} = \Sigma^{-1}(\vec{\mu}_1 - \vec{\mu}_0)$$

That is, the probability of an input **x** being in a class $y$ is purely a function of this linear combination of the known observations.

Fisher's linear discriminant is generalization of LDA, as it does not make some of the assumptions of LDA such as normally distributed classes or identical variances. The Fisher's linear discriminant method predicts the classes as follows:

Let the mean of two classes is $\vec{\mu}_0, \vec{\mu}_1$ and covariance is $\Sigma_{y=0}, \Sigma_{y=1}$, the linear combination of features will have means $\vec{w}.\mu_{y=i}$ and variances $\vec{w}^T.\Sigma_{y=i}\vec{w}$ for $i=0,1$. The criterion for separation the two classes to be the ratio of the variance between the classes to the variance within the classes. Formally

$$S = \frac{(\vec{w}(\vec{\mu}_{y=1} - \vec{\mu}_{y=0}))^2}{\vec{w}^T(\Sigma_{y=0} + \Sigma_{y=1})\vec{w}}$$

$S$ is used as measure of class labeling. It can be shown that the maximum separation occurs when

$$\vec{w} = (\Sigma_{y=0} + \Sigma_{y=1})^{-1}(\vec{\mu}_{y=1} - \vec{\mu}_{y=0})$$

We used 1700 abstracts related to pancreatic cancer from PUBMED and classified them with the help of experts into three categories, namely, non-evidence, intervention and non-intervention. Experts identified 420(24.7%) intervention abstracts, 250(14.7%) non-intervention abstracts and all other abstracts as non-evidence based abstracts from the collection of 1700 pancreatic cancer abstracts.

Three sets of features are used to represent the medical documents namely, *Medical Phrases*, *Semantic Tags*, and *Medical Phrases + Semantic Tags*. *Medical Phrases* and *Semantic Tags* are extracted using MMTx program. It also performs stopword removal and stemming. The details about each feature set are given in Table II. In each document the features are assigned weights according to their frequency of occurrence in the document. The feature sets used for document classification along with the sample training data sets are made available on the web[1].

TABLE II
SET OF FEATURES FOR 1700 ABSTRACTS ON PANCREATIC CANCER

| Classifier | Medical Phrases | Semantic Tags | Medical phrases + Semantic Tags |
|---|---|---|---|
| No. of Features | 9260 | 345 | 9605 |

We compare each of the classification algorithms mentioned above on *Medical Phrases*, *Semantic Tags* and *Medical Phrases + Semantic Tags*. Though some times the classification accuracy of using medical phrases with semantic tags is less, it offers a great help in interpreting the document and classification of medical questions. We used 10-fold cross validation for evaluating the performance of the classification algorithms on all three feature sets. The obtained results are shown in Table III.

TABLE III
ACCURACY OF DOCUMENT CLASSIFICATION METHODS

| Classifier | Medical Phrases | Semantic Tags | Medical phrases + Semantic Tags |
|---|---|---|---|
| SVM | 87.5079% | 87.4706% | 87.7769% |
| KNN | 87.6471% | 86.1765% | 86.8824% |
| NAÏVE BAYES | 75.4706% | 57.7647% | 74.7647% |
| DECISION TREE | 84.6741% | 83.8824% | 84.7762% |
| LDA | 75.5904% | 74.5731% | 75.8249% |

*B. Question Classification*

The question processing is the vital component of any question answering system, which is used to determine what is being sought in the question. In General Question Answering Systems (GQAS), the structure of the question reveals the desired answer type directly. GQAS focus on *what, when, who, where* and *why* question words, which may have named entities, such as person, organization and location as answer types. Most of the existing GQAS depend on the syntactic structure or hand crafted rules, as shown below:

> What {is | are} <phrase_to_define>?
> What is the definition of <phrase_to_define>?
> Who {is | was | are | were} <person_name(s)>?

The hand crafted rules created for a specific domain cannot be used for other domains. The rules created for specific question ontology must be redesigned before being applied to different ontologies. To overcome these problems machine learning methods have been researched and applied successfully [26].

Crafting rules for question classification in biomedical domain is difficult due to the complex structure of the questions, which may include patient description, test finding and so on. Rule based classification of the clinician's queries based on *what, when, who, where* and *why* question words and structure is difficult. For example, the question [10] given below, contains patients description which increased the complexity of interpreting the answer type.

---

[1] http://cc.domaindlx.com/zahed/cliniqa/

> Q: Do patients surviving an AMI and experiencing transient or ongoing congestive heart failure (CHF) have reduced mortality ad morbidity when treated with an ACE inhibitor (ex. Ramipril)?

A medical question analysis by clinicians in [4-6,17] can be used for classifying the medical questions and identifying the question focus. These studies are based on the questions asked by medical students and clinicians during practice. As experts have analyzed them, they are sound enough to be used in an automated medical question answering system.

The questions whose answers incorporate clinical studies are considered as good questions, however, all other questions are considered as patient specific questions and cannot be answered [6,7]. We select a subset of questions as good questions from a set of 10 generic questions given in [5]. We identify the focus of each question with the help of experts. The set of questions and their focus is shown in Table IV.

TABLE IV
CLASSES FOR IDENTYFING QUESTION FOCUS

| Class Number | Generic Medical Questions | Question Focus |
| --- | --- | --- |
| 1 | What is the drug choice for condition X? | clinical drug, Pharmacologic substance |
| 2 | What is the dosage of drug X? | Laboratory or Test results, Sign or symptom |
| 3 | How should I treat or manage condition X? | Therapeutic or preventive procedure; Diagnostic procedure |
| 4 | Can drug X cause (adverse) finding Y? | Qualitative Concept |

Recently, the researchers have focused on machine learning techniques for classifying medical questions. Yu et al. [7] used different classifiers to classify a set of 200 questions to the medical question taxonomy defined in [6]. Their result shows that SVM's performance is better than the other methods with 59.5% accuracy when used in ladder approach, and less than 54% when multi-class flat categorization is used. This accuracy is far from acceptable for a real time question answering system.

In CLINIQA, we use two classifiers in question classification, first to classify answerable and unanswerable questions and second classifier to map the question to any of the four classes defined in Table IV used to identify the question focus. A set of 100 questions related to pancreatic cancer are classified with the help of experts into four categories listed in Table IV. The questions are distributed as: class 1 (40 questions), class 2 (28 questions), class 3 (16 questions) and class 4 (16 questions).

Similar sets of features are used to represent the clinical question, which are used for document classification, namely, *Medical Phrases*, *Semantic Tags*, and *Medical Phrases + Semantic Tags*. The detail about each feature set is given in Table V. In each query the features are assigned binary weights with respect to their presence or absence in the query. The feature sets used for question classification along with the sample training data sets are made available on the web[1].

TABLE V
SET OF FEATURES FOR 100 QUESTIONS ON PANCREATIC CANCER

| Classifier | Medical Phrases | Semantic Tags | Medical phrases + Semantic Tags |
|---|---|---|---|
| No. of Features | 317 | 78 | 395 |

For the first classifier, to classify the question into answerable and unanswerable questions we use SVM, which gave a classification accuracy of 96.49%. The obtained results indicate that SVM gives 64.47% accuracy. Therefore, we use it also as the second classifier in question classification module of the CLINIQA. The result of classifying the questions into four categories is shown in Table VI.

---

[1] http://cc.domaindlx.com/zahed/cliniqa/

TABLE VI
QUESTION CLASSIFICATION ACCURACY (FOR IDENTIFYING QUESTION FOCUS)

| Classifier | Medical Phrases | Semantic Tags | Medical Phrases + Semantic Tags |
|---|---|---|---|
| SVM | 66.67% | 58.35% | 64.47% |
| KNN | 50.05% | 53.67% | 52.76% |
| NAÏVE BAYES | 53.59% | 53.47% | 53.87% |
| DECISION TREE | 57.78% | 52.71% | 57.54% |
| LDA | 52.73% | 47.59% | 49.86% |

Once the question classification is performed the medical phrases and semantics are directed to the query formulation module.

*C. Query formulation*

Query formulation module receives the medical phrases and semantic tags from the question classification module. We use vector space model due to its simplicity and high performance in TREC conference evaluation [29]. At this stage, the query formulation module assigns weights to the query terms. Both the documents as well as the queries are represented as vectors of term weights. In a corpus, with total *n* index terms in an entire collection of documents, the document D and the query Q can be represented as follows:

$$D = (w_{d1}, w_{d2}, w_{d3}, ...., w_{dn})$$

$$Q = (w_{q1}, w_{q2}, w_{q3}, ...., w_{qn})$$

where $w_{di}, w_{qi}$ are 'term weights' assigned to different terms in documents and query respectively. The term weights represent the statistical properties of the terms in documents or queries. The term weighting strategies for document are given in document classification subsection of section *A*. We use binary weights for query term weights, as the terms in queries are rarely repeated and scaling them to some other factors does not show any significant improvement in the results.

The similarity between document and query is often measured using Cosine metric and is commonly used to represent the Retrieval Status Value (RSV) for the document. The documents with high RSV are more similar or relevant to the query terms. It can be calculated as follows:

$$Cosine(Q,D) = \frac{\sum_{i=1}^{n} w_{qi} w_{di}}{\sqrt{\sum_{i=1}^{n} w_{qi}^2} \times \sqrt{\sum_{i=1}^{n} w_{di}^2}}$$

Where Q is query vector, D is document vector, $w_{qi}$ is weight of $i^{th}$ Query term, and $w_{di}$ is the weight of $i^{th}$ term in the document. The numerator is inner product of document and query vectors, while the denominator is product of their lengths. The denominator is used to normalized the document and query vectors to unit vectors. The effect of normalization is discussed in section II.A.

As the term weights in $Document \times Term$ matrix are assigned using Normalized term frequency scheme, and binary weights are assigned to the query terms (normalizing then will not affect the result), the cosine measure for similarity can be modified and computed as follows:

$$Sim(Q,D) = \sum_{i=1}^{n} w_{qi} w_{di}$$

We use same similarity measure for semantic tags. Finally, the measured similarities on both medical phrases and semantics are given as input to the answer extraction module.

*D. Answer extraction*

The answer extraction module accepts two sets of similarity vectors, corresponding to medical phrases and semantics as input, and identifies a set of abstracts that fall above certain threshold. A consensus between two similarity vectors is drawn by a simple heuristic. As a first step, the common documents to both the vectors are selected. In second step, the similarity scores of both medical phrases and semantics are added to get a new similarity score. Here, the bias can be incorporated while adding two similarity measures. For example, the semantic tags represent the generic classes; therefore they can have higher bias. Currently, we use simple addition of two similarity scores for simplicity without any bias to the vectors. The top ten abstracts are selected as the potential candidates for answering the user's question. The selected candidates are given as input to the answer ranking module, which ranks the candidates and returns the most relevant abstract to user.

*E. Answer ranking*

The answer ranking module is responsible for the ranking of the candidate abstracts based on sentence scores. The scores are assigned to sentences using question focus obtained from question classification module. There are two advantages of using question focus: (1) ranking of the answers to assign the top rank to the most relevant abstract, and (2) elimination of the abstracts which are selected by the answer extraction module due to high frequency of the general question words.

We rank the answer based on the simple heuristics as discussed below. The method first sets the flag associated with each sentence of the abstract if it contains the question focus, otherwise, the flag is assigned a zero value. In the second step, the sentences with flags having values 1 are investigated for the percentage of matching question words in the sentence. The percentage of matching question words is then assigned as the score of the sentence. Finally, each sentence score in the abstract is added to obtain the abstract score.

In case of a tie between the abstract scores, an abstract which contains a sentence with maximum sentence score is given the priority. When there is a tie between the abstract scores and sentence score then one of the abstracts is arbitrarily selected for the better rank, and the other abstract for the subsequent rank.

The sentence and abstract scoring is as follows:

$$f_{i,j} = \begin{cases} 1 & \text{if the sentence contains question focus} \\ 0 & \text{otherwise} \end{cases}$$

*where i is the abstracts number and j is sentence number.*

$$line\_score_{i,j} = f_{i,j} \times \% \text{ of question terms occuring in the sentence}$$

$$abstract\_score_i = \sum_{j=1}^{m} line\_score_{i,j}$$

*where m is total number of lines in abstract i*

After ranking, the abstracts are presented to the user based on their rank. The results contain the abstracts with the sentences with maximum scores highlighted. The snapshot of the CLINIQA's user interface is shown in Fig. 5.

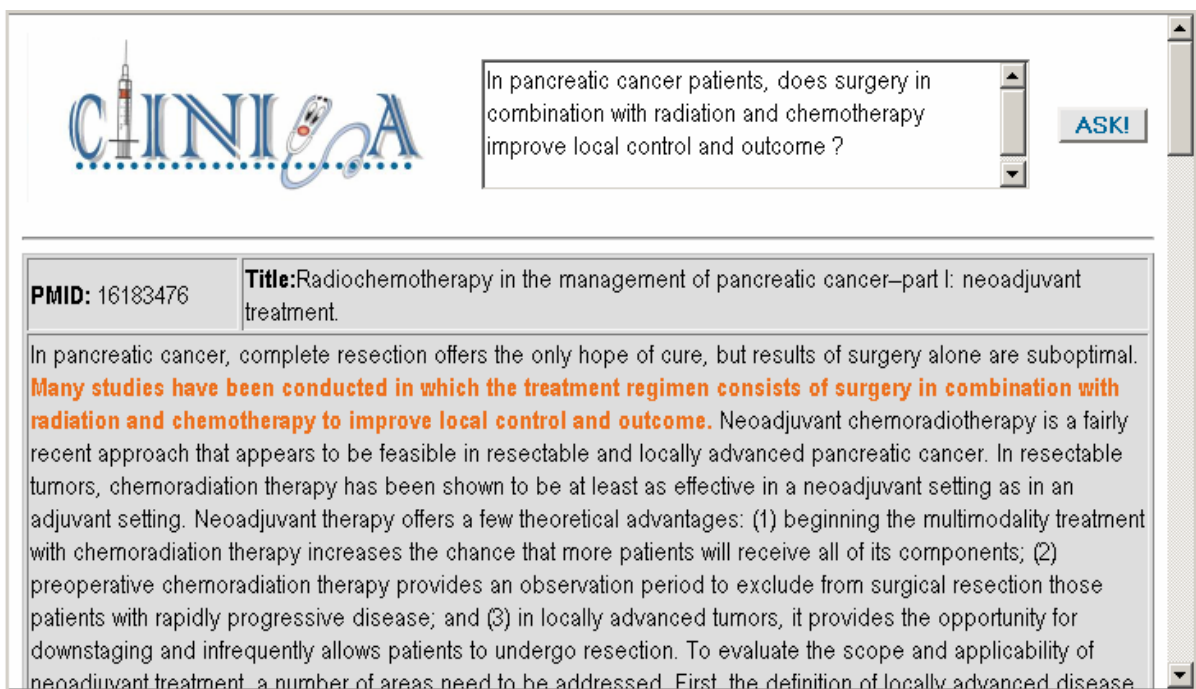

Fig. 5. CLINIQA user interface.

## III. Performance evaluation

In this section we evaluated CLINIQA based on mean reciprocal ranking and user effort measures and also compared the performance of the CLINIQA using different term weighting schemes. The evaluation results of three components, indexing and term weighting, document classification and question classification have been already presented in the previous section. Very little work [30,31] has been published focusing the evaluation methods for restricted domain question answering systems. Diekema et al. [30] described the factors other than TREC evaluation to assess the restricted domain question answering.

In the following section *A*, we discussed the evaluation criteria used to assess CLINIQA and using different weighting schemes. Section *B* describes the data set used for the experiments and in section *C* we present a comparative study of different information retrieval systems with respect to medical domain.

### A. Evaluation metrics

We used two metrics to evaluate the system, user effort and mean reciprocal ranking suggested in TREC [32] for the assessment of question answering systems.

Traditional information retrieval systems use the precision and recall to assess the retrieval performance. Precision is the percentage of retrieved documents that is relevant and recall is the percentage of relevant documents retrieved. The precision can be viewed as the measure of the effort user has to extend to find the relevant documents and recall measures how efficient the system is in locating the relevant documents. In question answering the recall can be defined as the percentage of questions answered correctly from the test set [33].

As we do not use web for extracting candidate answers the user effort function is slightly different from the one used in [33]. Let $A_1, A_2, A_3, \ldots, A_n$ be the $n$ abstracts selected by the answer ranking module and returned to user as the result, and $|A_i|$ represents the number of words in abstract $A_i$. If the answer is available in $i^{th}$ abstract then the user effort to reach the correct answer $|A_{ia}|$ is the number of word read before reaching the correct answer in abstract $A_i$. Then the new user effort calculation function is as follows:

$$Eff = \left(\sum_{i=1}^{n-1}|A_i|\right) + |A_{na}|$$

*where n is the abstract rank in which the answer is found.*

If the $n^{th}$ abstract has the answer then we assume that the user scans through $(n-1)$ abstracts completely, therefore the number of words in each of the $(n-1)$ abstracts are added to user effort. Finally the numbers of words till the answer sentence $|A_{na}|$ is added separately as shown in the effort calculation.

The performance of each question is computed by the reciprocal of the rank of the correct answer given by the system, which is defined as

$$RR = \frac{1}{rank_i} \; ; \; MRR = \frac{1}{n}\left(\sum_{i=1}^{n}\frac{1}{rank_i}\right)$$

*where n is the umber of questions*

Mean reciprocal rank is used to compute the overall performance of the question answering system and is defined above as *MRR*.

We evaluated CLINIQA on both user effort and MRR measures and the results are given in section *B*.

*B. Experimental evaluation*

To evaluate the system we used 1700 abstracts related to pancreatic cancer collected from PUBMED and 100 questions collected from clinicians and cancer discussion groups. The questions were distributed over the four classes, each of which identifies a unique question focus, as shown in section *2B*. Although the questions were selected from different sources we made sure that they were answerable using our document collection.

The document collection was then classified manually by the human experts into three categories, non evidence based, intervention and non-intervention based documents. The class distributions for document collection over the three classes are distributed as, 420 intervention documents, 250 non-intervention documents and the remaining documents were classified as non-evidence based documents.

As a first step, we evaluate the CLINIQA based on user effort with different weighting strategies such as TF, IDF, normalized term weights and ranking strategy. The experiment shows the superiority of the CLINIQA (with normalized weighting and abstract ranking strategy) over the other term weighting strategies without ranking. CLINIQA answered approximately 85%. Fig. 6 shows that the 36% of the questions were answered with user effort of 100 words using CLINIQA, whereas the 30% of the questions were answered by CLINIQA without ranking at the same user effort. In general, reading nearly 100 words take less than a minute by an average reader without any special reading skills. Approximately 40% of the questions are answered in the first abstract. Answers to the 50% of the questions is obtained in the first two abstracts, the remaining 35% of the questions answered at user effort ranging from 600 to 1200 words. An interesting study on reading capabilities reveal that an average person reads 280 words per minute [34] and another study reveal that physicians spend less than 2 minutes on average seeking an answer to a question [17]. The results discussed above indicate that CLINIQA offers a better solution for clinicians.

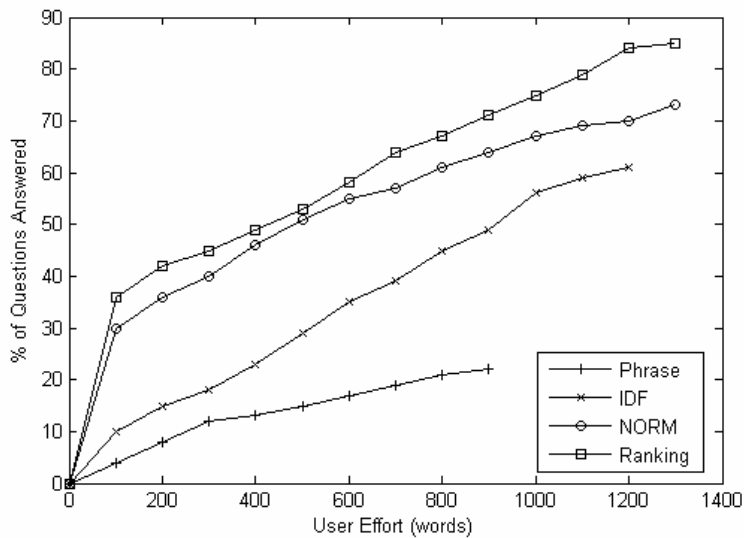

Fig. 6. Recall vs. User effort: when phrase frequency, inverse document frequency, normalized weights are used and normalized weights with abstract ranking strategy.

The evaluation of CLINIQA based on mean reciprocal ranking (MRR) is shown in Fig. 7. The results indicate that the ranking strategy helps in promoting the most relevant answer to the best rank, which in turn reduces the user effort, discussed above. CLINIQA ranks approximately 70% of the total questions answered at the top most rank whereas CLINIQA without ranking candidate answers approximately 60%.

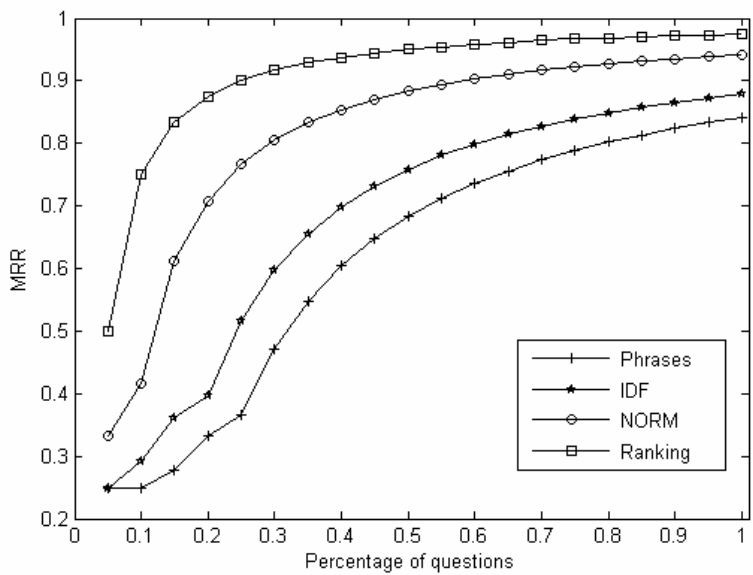

Fig. 7 The MRR results for CLINIQA with phrase frequency, inverse document frequency, normalized term weights, and normalized term weights with ranking.

IV. RELATED WORK

In recent years, the research in the area of medical question answering has received a great deal of interest. Most of the publications till date present an overview of the biomedical question answering system. One of early papers on biomedical question answering by Zweigenbaum [18] gives an overview of the system and reveals the different linguistic and knowledge sources available for the domain. Different strategies have been proposed for answering clinical questions such as, answering by role identification [10,19,20], and answering based on document structure [21].

The systems based on role identification accept queries in a PICO format, renowned in evidence based medicine [22]. In this format, a clinical question is represented by a set of four fields that correspond to the basic elements of the questions: P represents the description of patient, I represents intervention, C represents a comparison or control intervention and O represents the clinical outcome. Demner-Fushman et al. [20] worked on the extraction of PICO frame elements from MEDLINE abstracts. This cannot be considered as fully automated question answering system as format for questions is predefined, similar to any other query language. Another difficulty with the systems is that all the questions cannot be formatted as PICO [6].

Sang et al. [21] proposed an offline strategy for answering clinical questions. This was developed and evaluated for non-expert users. They did not consider diagnostic questions. Hence most of the questions were answerable from medical encyclopedias and FAQs. The system is simple and the strategies are not suitable for evidence-based medicine.

Significant contribution to individual components of the biomedical question answering has been done in recent years. A very effective and useful analysis of biomedical questions is given in [4-6,17] by the physicians. In [4,5,17], Ely et al., presented the significance of the medical question answering system, and proposed a list of generalized medical questions based on a survey conducted on medical students and practitioners. They also presented good list of obstacles in answering medical questions, from a physician's point of view. Interestingly, to the best of our knowledge, no biomedical question answering system has used the analysis presented in [4-6,17]. Recently, an effort is made by Yu et al. [5,6] to classify the medical questions statistically based on the generic classification.

The other components on which the whole MQAS depends are domain specific resources and knowledgebase. Domain specific resources such as dictionaries and thesauri are used as the major tools in general question answering systems. WordNet is a good example for such a system. The UMLS [2] developed by National Library of Medicine (NLM) provides a rich knowledge of medical terms and their semantics. UMLS contain three knowledge sources: Metathesaurus, Semantic Network and Specialist Lexicon.

The MEDLINE database is one of the most reliable sources for knowledgebase to answer the medical questions and is often used by the clinicians [23]. Studies have shown that the existing systems for searching MEDLINE are often inadequate and unable to supply clinically-relevant answers in a timely manner [24].

To the best of our knowledge CLINIQA is the first machine intelligence based clinical question answering system, whose components are thoroughly evaluated to arrive at a highly accurate system. It classifies the questions based on generic classification of medical questions proposed in [4-6,17]. It makes use of UMLS for interpreting the user question and indexing medical documents. Finally, the PUBMED abstracts are used to extract and locate the most relevant answers. Further more, the architecture is robust and can used with any other medical literature knowledgebase related any other disease. We used UMLS for interpreting the medical documents and UMLS is designed for all kind of medical literature made CLINIQA robust for any medical knowledge source and disease.

## V. Conclusion

In this paper, we presented a novel implementation of machine learning based clinical question answering system. In this effort we tried to assist clinicians in evidence based information retrieval, by using authenticated and proved medical literature resources like PUBMED. The best machine learning tools were employed for the classification of document and questions. The system relies on the clinicians' study for question analysis. Effective answer ranking strategies are employed to reduce the user effort. CLINIQA is rigorously evaluated from the user's point of view and the performance point of view of supervised machine learning techniques. The results indicate that CLINIQA helps clinicians to obtain answers to approximately 50% of questions with a user effort of approximately 1 minute. These results support CLINIQA for clinicians' use in daily practice.

## References

[1]  URL: http://www.ncbi.nlm.nih.gov/
[2]  URL: http://www.nlm.nih.gov/research/umls/
[3]  URL: http://mmtx.nlm.nih.gov/
[4]  G. R. Bergus, C. S. Randall, S.D Sinift, and D. M. Rosenthal, "Does the structure of clinical questions affect the outcome of the curbside consultations with specialty colleagues?," *Arc Fam Med*, vol. 9, 2000, pp. 541-547.
[5]  J. Ely, J. Osheroff, P. Gorman, M. Ebell, M. Chambliss, E. Pifer, and P. Stavri, "A taxonomy of generic clinical questions: classification study," *British Medical Journal* vol. 321, 2000, pp. 429-432.
[6]  J. Ely, J. Osheroff, M. Ebell, M. Chambliss, D. Vinson, J. Stevermer, and E. Pifer, "Obstacles to answering doctors' questions about patient care with evidence: qualitative study," *British Medical Journal*, vol. 324. 2002, pp. 710-713.
[7]  H. Yu, C. Sable, H. R. Zhu, "Classifying medical questions based on evidence taxonomy," unpublished.
[8]  S. Schultz, M. Honeck, and H. Hahn, "Biomedical text retrieval in languages with complex morphology," in proceedings of the *Workshop on Natural Language Processing in the Biomedical domain*, July 2002, pp. 61-68.
[9]  Y. Song, S. Kim, and H. Rim, "Terminology Indexing and Reweighting methods for Biomedical Text Retrieval," In Proceedings of the *SIGIR'04 Workshop on Search and Discovery in Bioinformatics*, ACM, Sheffield, UK, 2004.